\documentclass[11pt,a4paper]{article}
\usepackage{acl2015}
\usepackage{times}
\usepackage{url}
\usepackage[small]{caption}
\usepackage{latexsym}
\usepackage{color,multirow,rotating}
\usepackage[table,xcdraw]{xcolor}
\usepackage{booktabs}
\usepackage{multirow}
\usepackage{amsmath, amsthm, amssymb, amsfonts}

\makeatletter
\newcommand{\@BIBLABEL}{\@emptybiblabel}
\newcommand{\@emptybiblabel}[1]{}
\newcommand\footnoteref[1]{\protected@xdef\@thefnmark{\ref{#1}}\@footnotemark}

\makeatother
\usepackage{url}

\usepackage{xspace}
\newcommand{\VAndL}{vision \& language\xspace}
\newcolumntype{R}[1]{>{\begin{sideways}}p{#1}<{\end{sideways}}}
\title{A Survey of Current Datasets for Vision and Language Research}

\newcommand{\JHUAff}[0]{\ensuremath{1}\xspace}
\newcommand{\URAff}[0]{\ensuremath{2}\xspace}
\newcommand{\CMUAff}[0]{\ensuremath{3}\xspace}
\newcommand{\MSRAff}[0]{\ensuremath{4}\xspace}

\author{{Francis Ferraro{$^{\JHUAff}$\thanks{\ \ F.F. and N.M. contributed equally to this work.}}\xspace~}, {Nasrin Mostafazadeh{$^{\URAff}$\footnotemark[1]}\xspace~}, Ting-Hao (Kenneth) Huang{$^\CMUAff$},\\
{\bf Lucy Vanderwende$^\MSRAff$, Jacob Devlin$^\MSRAff$,
Michel Galley$^\MSRAff$, Margaret Mitchell$^\MSRAff$} \\
\\
{\bf Microsoft Research} \\
\\
{\small $\JHUAff$} Johns Hopkins University, 
{\small $\URAff$} University of Rochester, 
{\small $\CMUAff$} Carnegie Mellon University, \\
{\small $\MSRAff$} Corresponding authors: \{lucyv,jdevlin,mgalley,memitc\}@microsoft.com
}

\begin{document}

\maketitle

\begin{abstract}
Integrating vision and language has long been a dream in work on artificial intelligence (AI). In the past two years, we have witnessed an explosion of work that brings together vision and language from images to videos and beyond. The available corpora have played a crucial role in advancing this area of research. In this paper,  we propose a set of quality metrics for evaluating and analyzing the vision \& language datasets and categorize them accordingly. Our analyses show that the most recent datasets have been using more complex language and more abstract concepts, however, there are different strengths and weaknesses in each.
\end{abstract}


\section{Introduction}

Bringing together language and vision in one intelligent system has long been an ambition in AI research, beginning with SHRDLU as one of the first vision-language integration systems \cite{winograd1972SHRDLU} and continuing with more recent attempts on conversational robots grounded in the visual world \cite{kollar2013groundedRobotLearning,cantrell2010HRI,matuszekgroundedattribute2012,kruijff2007robotDialogue,roy2003conversationalRobots}. In the past few years, an influx of new, large \VAndL corpora, alongside dramatic advances in vision research, has sparked renewed interest in connecting vision and language.  Vision \& language corpora now provide alignments between visual content that can be recognized with Computer Vision (CV) algorithms and language that can be understood and generated using Natural Language Processing techniques.

Fueled in part by the newly emerging data, research that blends techniques in vision and in language has increased at an incredible rate.  In just the past year, recent work has proposed methods for image and video captioning \cite{fang2014captionsToConcepts,donahue2014RCNN,venugopalan2015videoNN}, summarization \cite{gunhee2015blogStory}, reference \cite{kazemzadeh2014referitgame}, and question answering \cite{antol2015vqa,gao2015mQA}, to name just a few. The newly crafted large-scale \VAndL datasets have played a crucial role in defining this research, serving as a foundation for training/testing and helping to set benchmarks for measuring system performance.


Crowdsourcing and large image collections such as those provided by Flickr\footnote{~\url{http://www.flickr.com}} have made it possible for researchers to propose methods for vision and language tasks alongside an accompanying dataset.  However, as more and more datasets have emerged in this space, it has become unclear how different methods generalize beyond the datasets they are evaluated on, and what data may be useful for moving the field beyond a single task, towards solving larger AI problems.

In this paper, we take a step back to document this moment in time, making a record of the major available corpora that are driving the field. We provide a quantitative analysis of each of these corpora in order to understand the characteristics of each, and how they compare to one another. The quality of a dataset must be measured and compared to related datasets, as low quality data may distort an entire subfield. We propose a set of criteria for analyzing, evaluating and comparing the quality of \VAndL datasets against each other. Knowing the details of a dataset compared to similar datasets allows researchers to define more precisely what task(s) they are trying to solve, and select the dataset(s) best suited to their goals, while being aware of the implications and biases the datasets could impose on a task. 

We categorize the available datasets into three major classes and evaluate them against these criteria. The datasets we present here were chosen because they are all available to the community and cover the data that has been created to support the recent focus on image captioning work. More importantly, we provide an evolving website\footnote{\label{url-footnote}\url{http://visionandlanguage.net}} containing pointers and references to many more vision-to-language datasets, which we believe will be valuable in unifying the quickly expanding research tasks in language and vision.

\section{Quality Criteria for Language \& Vision Datasets}\label{sec:criteria}

The quality of a dataset is highly dependent on the sampling and scraping techniques used early in the data collection process. However, the content of datasets can play a major role in narrowing the focus of the field. 
Datasets are affected by both {\it reporting bias} \cite{gordon2013reportingBias}, where the frequency with which people write about actions, events, or states does not directly reflect real-world frequencies of those phenomena
; they are also affected by {\it photographer's bias} \cite{torralba2011biasInVision}, where photographs are somewhat predictable within a given domain. This suggests that new datasets may be useful towards the larger AI goal if provided alongside a set of quantitative metrics that show how they compare against similar corpora, as well as more general ``background'' corpora.  Such metrics can be used as indicators of dataset bias and language richness.  At a higher level, we argue that clearly defined metrics are necessary to provide quantitative measurements of how a new dataset compares to previous work. This helps clarify and benchmark how research is progressing towards a broader AI goal as more and more data comes into play. 

In this section, we propose a set of such metrics that characterize \VAndL datasets.  We focus on methods to measure {\it language quality} that can be used across several corpora.  We also briefly examine metrics for {\it vision quality}. 
We evaluate several recent datasets based on all proposed metrics in Section~\ref{sec:analysis}, with results reported in Tables~\ref{tab:stats}, \ref{tab:ppl}, and Figure~\ref{fig:pos_distr}.

\subsection{Language Quality} 
We define the following 
criteria for evaluating the captions or instructions of the datasets:\\
$\bullet$ \textbf{Vocabulary Size} (\textit{\#vocab}), the number of unique vocabulary words.\\
$\bullet$ \textbf{Syntactic Complexity} (\textit{Frazier}, \textit{Yngve}) measures the amount of embedding/branching in a sentence's syntax. We report mean Yngve \cite{yngve60syntax} and Frazier measurements \cite{frazier85syntax}; each provides a different counting on the number of nodes in the phrase markers of syntactic trees.\\
$\bullet$ \textbf{Part of Speech Distribution} measures the distribution of nouns, verbs, adjectives, and other parts of speech.\\
$\bullet$ \textbf{Abstract:Concrete Ratio} ($\#\textit{Conc}$, $\#\textit{Abs}$, $\%\textit{Abs}$) indicates the range of visual and non-visual concepts the dataset covers. Abstract terms are ideas or concepts, such as `love' or `think' and concrete terms are all the objects or events that are mainly available to the senses. For this purpose, we use a list of most common abstract terms in English \cite{vanderwende2015AMR}, and define concrete terms as all other words except for a small set of function words.\\ 
$\bullet$ \textbf{Average Sentence Length} (\textit{Sent Len.}) shows how rich and descriptive the sentences are.\\
$\bullet$ \textbf{Perplexity} provides a measure of data skew by measuring how expected sentences are from one corpus according to a model trained on another corpus. We analyze perplexity (\textit{Ppl}) for each dataset against a 5-gram language model learned on a generic 30B words English dataset. We further analyze pair-wise perplexity of datasets against each other in Section~\ref{sec:analysis}.

\subsection{Vision Quality}
Our focus in this survey is mainly on language, however, the characteristics of images or videos and their corresponding annotations is as important in \VAndL research. The quality of vision in a dataset can be characterized in part by the variety of visual subjects and scenes provided, as well as the richness of the annotations (e.g., segmentation using bounding boxes (\textit{BB}) or visual dependencies between boxes). Moreover, a vision corpus can use abstract or real images (\textit{Abs/Real}).


\section{The Available Datasets}\label{sec:datasets}
We group a representative set of available datasets based on their content. For a complete list of datasets and their descriptions, please refer to the supplementary website.\footnoteref{url-footnote}

\subsection{Captioned Images}
Several recent \VAndL datasets provide one or multiple captions per image. The captions of these datasets are either the original photo title and descriptions provided by online users \cite{Ordonez:2011:im2text,thomee2015yfcc100m}, or the captions generated by crowd workers for existing images. The former datasets tend to be larger in size and contain more contextual descriptions.
\subsubsection{User-generated Captions}

$\bullet$ \textbf{SBU Captioned Photo Dataset} \cite{Ordonez:2011:im2text} contains 1 million images with original user generated captions, collected in the wild by systematic querying of Flickr. This dataset is collected by querying Flickr for specific terms such as objects and actions and then filtered images with descriptions longer than certain mean length.\\ 
$\bullet$ \textbf{D\'{e}j\`{a} Images Dataset} \cite{chen2015deja} consists of 180K unique user-generated captions associated with 4M Flickr images, where one caption is aligned with multiple images. This dataset was collected by querying Flickr for 693 high frequency nouns, then further filtered to have at least one verb and be judged as ``good'' captions by workers on Amazon's Mechanical Turk (Turkers).

\begin{table*}[htbp]
\centering
\resizebox{\textwidth}{!}{
\begin{tabular}{ccrrrrrrrrrccc}
\hline
 &  & \multicolumn{2}{c}{{\bf Size(k)}} & \multicolumn{8}{c}{{\bf Language}} & \multicolumn{2}{c}{{\bf Vision}} \\ \hline
 & {\bf Dataset} & \multicolumn{1}{c}{{\bf Img}} & \multicolumn{1}{c}{{\bf Txt}} & \multicolumn{1}{c}{{\bf Frazier}} & \multicolumn{1}{l}{{\bf Yngve}} & \multicolumn{1}{c}{{\bf \begin{tabular}[c]{@{}c@{}}Vocab\\ Size (k)\end{tabular}}} & \multicolumn{1}{c}{{\bf \begin{tabular}[c]{@{}c@{}}Sent\\ Len.\end{tabular}}} & \multicolumn{1}{c}{{\bf \#Conc}} & \multicolumn{1}{c}{{\bf \#Abs}} & \multicolumn{1}{c}{{\bf \%Abs}} & {\bf Ppl} & {\bf \begin{tabular}[c]{@{}c@{}}(A)bs/\\ (R)eal\end{tabular}} & {\bf BB} \\ \hline
 {\bf Balanced}
& {\bf Brown} & - & 52 & 18.5 & 77.21 & 47.7 & 20.82 & 40411 & 7264 & 15.24\% & 194  & - & - \\ \hline
\multirow{3}{*}{{\bf User-Gen}} 
& {\bf SBU} & 1000 & 1000 & 9.70 & 26.03 & 254.6 & 13.29 & 243940 & 9495 & 3.74\% & 346 & R & - \\ \cline{2-14}
& {\bf Deja} & 4000 & 180 & 4.13 & 4.71 & 38.3 & 4.10 & 34581 & 3714 & 9.70\% & 184 & R & - \\ \hline
\multirow{3}{*}{{\bf \begin{tabular}[c]{@{}c@{}}Crowd-\\ sourced\end{tabular}}}
& {\bf Pascal} & 1 & 5 & 8.03 & 25.78 & 3.4 & 10.78 & 2741 & 591 & 17.74\% & 123 & R & - \\ \cline{2-14}
& {\bf Flickr30K} & 32 & 159 & 9.50 & 27.00 & 20.3 & 12.98 & 17214 & 3033 & 14.98\% & 118 & R & - \\ \cline{2-14}
 & {\bf COCO} & 328 & 2500 & 9.11 & 24.92 & 24.9 & 11.30 & 21607 & 3218 & 12.96\% & 121 & R & Y \\ \cline{2-14} 
 & {\bf Clipart} & 10 & 60 & 6.50 & 12.24 & 2.7 & 7.18 & 2202 & 482 & 17.96\% & 126 & A & Y \\ \hline
{\bf Video} & {\bf VDC} & 2 & 85 & 6.71 & 15.18 & 13.6 & 7.97 & 11795 & 1741 & 12.86\% & 148 & R & - \\ \hline
\multirow{3}{*}{{\bf Beyond}} & {\bf VQA} & 10 & 330 & 6.50 & 14.00 & 6.2 & 7.58 & 5019 & 1194 & 19.22\% & 113 & A/R & - \\ \cline{2-14} 
 & {\bf CQA} & 123 & 118 & 9.69 & 11.18 & 10.2 & 8.65 & 8501 & 1636 & 16.14\% & 199 & R & Y \\ \cline{2-14} 
 & {\bf VML} & 11 & 360 & 6.83 & 12.72 & 11.2 & 7.56 & 9220 & 1914 & 17.19\% & 110 & R & Y \\ \hline
\end{tabular}
}
\caption{Summary of statistics and quality metrics of a sample set of major datasets.
For Brown, we report Frazier and Yngve scores on automatically acquired parses, but we also compute them for the 24K sentences with gold parses: in this setting, the mean Frazier score is 15.26 while the mean Yngve score is 58.48.}
\label{tab:stats}
\end{table*}


	
\subsubsection{Crowd-sourced Captions}

$\bullet$ \textbf{UIUC Pascal Dataset} \cite{farhadi2010pascal} is probably one of the first datasets aligning images with captions. Pascal dataset contains 1,000 images with 5 sentences per image.\\
$\bullet$ \textbf{Flickr 30K Images} \cite{young2014image} extends previous Flickr datasets \cite{rashtchian2010flickr8k}, and includes 158,915 crowd-sourced captions that describe 31,783 images of people involved in everyday activities and events.\\
$\bullet$ \textbf{Microsoft COCO Dataset (MS COCO)} \cite{coco2014} includes complex everyday scenes with common objects in naturally occurring contexts. Objects in the scene are labeled using per-instance segmentations. In total, this dataset contains photos of 91 basic object types with 2.5 million labeled instances in 328k images, each paired with 5 captions. This dataset gave rise to the CVPR 2015 image captioning challenge and is continuing to be a benchmark for comparing various aspects of vision and language research.\\
$\bullet$ \textbf{Abstract Scenes Dataset (Clipart)} \cite{zitnick2013abstractScenes} was created with the goal of representing real-world scenes with clipart to study scene semantics isolated from object recognition and segmentation issues in image processing.  This removes the burden of low-level vision tasks.  This dataset contains 10,020 images of children playing outdoors associated with total 60,396 descriptions.

\subsubsection{Captions of Densely Labeled Images}

Existing caption datasets provide images paired with captions, but such brief image descriptions capture only a subset of the content in each image.  Measuring the magnitude of the reporting bias inherent in such descriptions helps us to understand the discrepancy between what we can learn for the specific task of image captioning versus what we can learn more generally from the photographs people take.  
One dataset useful to this end provides image annotation for content selection: \\
$\bullet$ \textbf{Microsoft Research Dense Visual Annotation Corpus} \cite{yatskar2014dense} provides a set of 500 images from the Flickr 8K dataset \cite{rashtchian2010flickr8k} that are densely labeled with 100,000 textual labels, with bounding boxes and facets annotated for each object.  This approximates ``gold standard'' visual recognition.

To get a rough estimate of the reporting bias in image captioning, we determined the percentage of top-level objects\footnote{This visual annotation consists of a two-level hierarchy, where multiple Turkers enumerated and located objects and stuff in each image, and these objects were then further labeled with finer-grained object information ({\it Has} attributes).}
that are mentioned in the captions for this dataset out of all the objects that are annotated.  
Of the average 8.04 available top-level objects in the image, each of the captions only reports an average of 2.7 of these objects.\footnote{We did not use an external synonym or paraphrasing resource to perform the matching between labels and captions, as the dataset itself provides paraphrases for each object: each object is labeled by multiple Turkers, who labeled {\it Isa} relations (e.g., ``eagle'' is a ``bird'').} 
A more detailed analysis of reporting bias is beyond the scope of this paper, but we found that many of the biases (e.g., people selection) found with abstract scenes \cite{zitnick2013abstractScenes} are also present with photos.

\subsection{Video Description and Instruction}
Video datasets aligned with descriptions \cite{chen2010sportcaster,rohrbach12cvpr,regneri2013groundingActions,naim2015instructionsVideo,malmaud2015cooking} generally represent  limited domains and small lexicons, which is due to the fact that video processing and understanding is a very compute-intensive task. Available datasets include: 
\\
$\bullet$ \textbf{Short Videos Described with Sentences} \cite{Yu2013groundedVideo} includes 61 video clips (each 3–5 seconds in length, filmed in three different outdoor environments), showing multiple simultaneous events between a subset of four objects: a person, a backpack, a chair, and a trash-can. Each video was manually annotated (with very restricted grammar and lexicon) with several sentences describing what occurs in the video.\\
$\bullet$ \textbf{Microsoft Research Video Description Corpus (MS VDC)} \cite{Chen2011paraphraseVideos} contains parallel descriptions (85,550 English ones) of 2,089 short video snippets (10-25 seconds long). The descriptions are one sentence summaries about the actions or events in the video as described by Amazon Turkers. In this dataset,  both paraphrase and bilingual alternatives are captured, hence, the dataset can be useful translation, paraphrasing, and video description purposes.

\subsection{Beyond Visual Description}
Recent work has demonstrated that n-gram language modeling paired with scene-level understanding of an image trained on large enough datasets can result in  reasonable automatically generated captions \cite{fang2014captionsToConcepts,donahue2014RCNN}. Some works have proposed to step beyond description generation, towards deeper AI tasks such as question answering \cite{ren2015imageqa,malinowski2014sceneQA}. We present two of these attempts below: 
\\
$\bullet$ \textbf{Visual Madlibs Dataset (VML)} \cite{yu2015vml} is a subset of 10,783 images from the MS COCO dataset which aims to go beyond describing which objects are in the image. For a given image, three Amazon Turkers were prompted to complete one of 12 fill-in-the-blank template questions, such as `when I look at this picture, I~feel --', selected automatically based on the image content. This dataset contains a total of 360,001 MadLib question and answers.\\
$\bullet$ \textbf{Visual Question Answering (VQA) Dataset} \cite{antol2015vqa} is created for the task of open-ended VQA, where a system can be presented with an image and a free-form natural-language question (e.g., `how many people are in the photo?'), and should be able to answer the question. This dataset contains both real images and abstract scenes, paired with questions and answers. Real images include 123,285 images from MS COCO dataset, and 10,000 clip-art abstract scenes, made up from 20 `paperdoll' human models with adjustable limbs and over 100 objects and 31 animals. Amazon Turkers were prompted to create `interesting' questions,  resulting in 215,150 questions and 430,920 answers.\\
$\bullet$ \textbf{Toronto COCO-QA Dataset (CQA)} \cite{ren2015imageqa} is also a visual question answering dataset, where the questions are automatically generated from image captions of MS COCO dataset. This dataset has a total of 123,287 images with 117,684 questions with one-word answer about objects, numbers, colors, or locations.

\begin{table*}[t!]
\small
\centering
\begin{tabular}{@{}rrrrrrrrrr@{}} \hline
\toprule
                    &{\bf Brown} &{\bf Clipart}         &{\bf Coco}         &{\bf Flickr30K}        &{\bf CQA}      &{\bf VDC}          &{\bf VQA} &{\bf Pascal} &{\bf SBU}\\ 
\midrule
{\bf Brown}         &237.1   &99.6   &560.8    &405.0   &354.039   &187.3   &126.5 &47.8 &621.5\\
{\bf Clipart}       &233.6   &11.2               &117.4              &109.4              &210.8              &82.5               &114.7 &28.7 &130.6\\
{\bf Coco}      &274.6     &59.2               &36.2               &75.3               &137.0              &87.1               &236.9 &39.3 &111.0\\
{\bf Flickr30K}  &247.8      &78.5               &54.3               &37.8               &181.5              &72.1               &192.2 &39.9 &125.0\\
{\bf CQA}        &489.4   &186.1              &137.0              &244.5              &33.8               &259.0              &72.1 &74.9 &200.1\\
{\bf VDC}        &200.5   &52.4               &61.5               &51.1               &289.9              &30.0               &180.1 &28.7 &154.5\\
{\bf VQA}        &425.9   &368.8              &366.8              &665.8              &317.7              &455.0              &19.6 &119.3 &281.0\\
{\bf Pascal}     &265.2      &64.5              &43.2          &63.4          &174.2           &83.0        &228.2 &36.0 &105.3 \\
{\bf SBU}        &473.9   &107.1           &346.4           & 344.0          &328.5        &230.7            &194.3 &78.2 &119.8\\

\midrule
{\it \#vocab} &{\it 14.0k} & {\it 1.1k} & {\it 13k} & {\it 9.4k} & {\it 5.3k} & {\it 4.9k} & {\it 1.4k} & {\it 1.0k} & {\it 65.1k}\\
\bottomrule
\end{tabular}
\caption{Perplexities across corpora, where rows represent test sets (20k sentences) and columns training sets (remaining sentences). To make perplexities comparable, we used the same vocabulary frequency cutoff of 3. All models are 5-grams.}
\label{tab:ppl}
\end{table*}

\section{Analysis}\label{sec:analysis}
\begin{figure}[t]
\centering
\includegraphics[scale=.44]{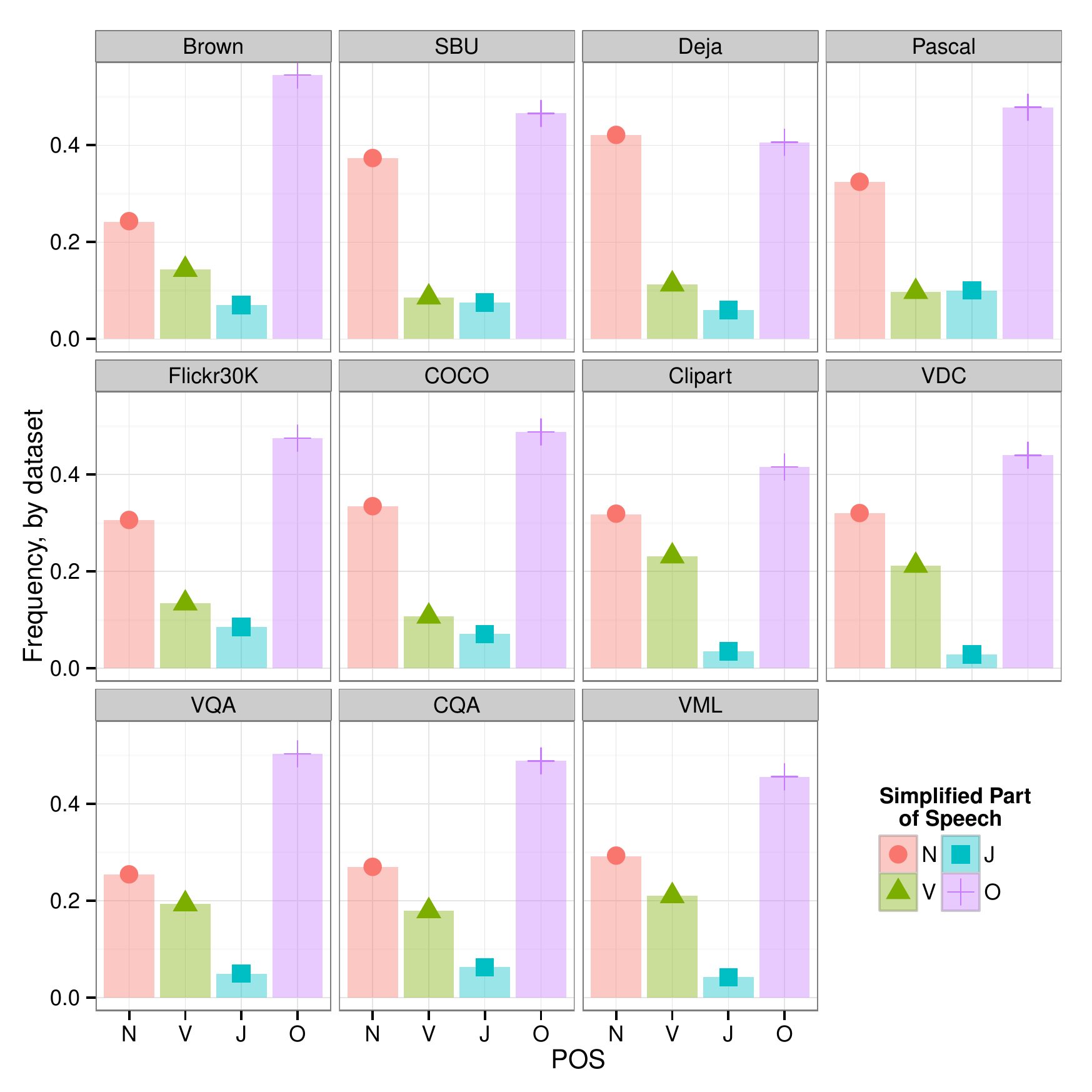}
\caption{Simplified part-of-speech distributions for the eight datasets.
We include the POS tags from the balanced Brown corpus  \cite{BrownCorpus}  to contextualize any very shallow syntactic biases.
We mapped all nouns to ``N,'' all verbs to ``V,'' all adjectives to ``J'' and all other POS tags to ``O.''
}
\label{fig:pos_distr}
\end{figure}

We analyze the datasets introduced in Section \ref{sec:datasets} according to the metrics defined in Section \ref{sec:criteria}, using the Stanford CoreNLP suite to acquire parses and part-of-speech tags \cite{manning-corenlp-2014}. We also include the Brown corpus \cite{francis1979brown,BrownCorpus} as a reference point. We find evidence that the VQA dataset captures more abstract concepts than other datasets, with almost 20\% of the words found in our abstract concept resource. The Deja corpus has the least number of abstract concepts, followed by COCO and VDC.  This reflects differences in collecting the various corpora:  For example, the Deja corpus was collected to find {\it specifically} visual phrases that can be used to describe multiple images.  This corpus also has the most syntactically simple phrases, as measured by both Frazier and Yngve; this is likely caused by the phrases needing to be general enough to capture multiple images.

The most syntactically complex sentences are found in the Flickr30K, COCO and CQA datasets.  However, the CQA dataset suffers from a high perplexity against a background corpus relative to the other datasets, at odds with relatively short sentence lengths.  This suggests that the automatic caption-to-question conversion may be creating unexpectedly complex sentences that are less reflective of general language usage.  In contrast, the COCO and Flickr30K dataset's relatively high syntactic complexity is in line with their relatively high sentence length.

Table~\ref{tab:ppl} illustrates further similarities between datasets, and a more fine-grained use of perplexity to measure the usefulness of a given training set for predicting words of a given test set. 
Some datasets such as COCO, Flickr30K, and Clipart are generally more useful as out-domain data compared to the QA datasets. Test sets for VQA and CQA are quite idiosyncratic and yield poor perplexity unless trained on in-domain data. As shown in Figure \ref{fig:pos_distr}, the COCO dataset is balanced across POS tags most similarly to the balanced Brown corpus  \cite{BrownCorpus}.  The Clipart dataset provides the highest proportion of verbs, which often correspond to actions/poses in vision research, while the Flickr30K corpus provides the most nouns, which often correspond to object/stuff categories in vision research.

We emphasize here that the distinction between a qualitatively good or bad dataset is task dependent. Therefore, all these metrics and the obtained results provide the researchers with an objective set of criteria so that they can make the decision whether a dataset is suitable to a particular task. 
\section{Conclusion}
We detail the recent growth of \VAndL corpora and compare and contrast several recently released large datasets.  We argue that newly introduced corpora may measure how they compare to similar datasets by measuring {\it perplexity}, {\it syntactic complexity}, {\it abstract:concrete} word ratios, among other metrics.  By leveraging such metrics and comparing across corpora, research can be sensitive to how datasets are biased in different directions, and define new corpora accordingly.

\paragraph*{Acknowledgements}
We thank the three anonymous reviewers for their feedback, and Benjamin Van Durme for discussions on reporting bias.

\bibliographystyle{acl}

\bibliography{bib}

\end{document}